\documentclass{article} 
\usepackage[left=2cm, right=2cm]{geometry}

\title{Causal Layering via Conditional Entropy}

\usepackage{amsmath}
\usepackage{graphicx}
\usepackage{amsthm}
\usepackage{amsfonts}
\newtheorem{theoremsepcount}{Theorem}

\newtheorem{lemmasepcount}{Lemma}
\newtheorem{definitionsepcount}{Definition}

\newtheorem{assumptioncollection}{Assumption Collection}
\usepackage{tikz}
\usetikzlibrary{decorations.pathreplacing}
\usetikzlibrary{fadings}
\usepackage{xcolor}
\usepackage{caption}
\usepackage{subcaption}
\usepackage{cancel}
\usepackage{authblk}
\usepackage[sort&compress,square,comma,authoryear]{natbib}

\DeclareMathOperator*{\argmax}{arg\,max}
\DeclareMathOperator*{\argmin}{arg\,min}

\usepackage{mathrsfs}
\newcommand{\indep}{\perp \!\!\! \perp}
\usepackage{algorithm, algorithmic}
\usepackage{enumitem}
\floatname{algorithm}{Algorithm}

\newcommand{\ASSUMPTIONS}{\item[\algorithmicassumptions]}
\newcommand{\algorithmicassumptions}{\textbf{Assumptions:}}
\usepackage{caption}

\author{Itai Feigenbaum\thanks{The author would like to thank Yu Bai for fruitful discussions.}}
\author{Devansh Arpit}
\author{Huan Wang}
\author{Shelby Heinecke}
\author{Juan Carlos Niebles}
\author{Weiran Yao}
\author{Caiming Xiong}
\author{Silvio Savarese}
\affil{Salesforce AI Research}
\date{}
\begin{document}
\maketitle

\begin{abstract}%
Causal discovery aims to recover information about an unobserved causal graph from the observable data it generates. Layerings are orderings of the variables which place causes before effects. In this paper, we provide ways to recover layerings of a graph by accessing the data via a conditional entropy oracle, when distributions are discrete. Our algorithms work by repeatedly removing sources or sinks from the graph. Under appropriate assumptions and conditioning, we can separate the sources or sinks from the remainder of the nodes by comparing their conditional entropy to the unconditional entropy of their noise. Our algorithms are provably correct and run in worst-case quadratic time. The main assumptions are faithfulness and injective noise, and either known noise entropies or weakly monotonically increasing noise entropies along directed paths. In addition, we require one of either a very mild extension of faithfulness, or strictly monotonically increasing noise entropies, or expanding noise injectivity to include an additional single argument in the structural functions.
\end{abstract}


\section{Introduction}

In the field of causality, data is generated by a causal graph. The purpose of causal discovery is to recover information about an unobserved causal graph via the observed data. One important task in causal discovery is the recovery of a topological ordering of the underlying graph, which in the context of causality is an ordering of the nodes which places causes before effects. Other than the importance of such ordering in its own right, it is also highly useful for discovering the full graph \citep{teyssier2005ordering}. A topological ordering which doesn't unnecessarily break all ties is called a {\it layering} \citep{tamassia2013handbook}.

Given a graph, repeated removal of sources or sinks yields a layering: we denote these algorithms as {\it repeatedSOUrceRemoval} (SOUR) and {\it repeatedSInkRemoval} (SIR), both are simple and probably known variants of Kahn's Algorithm \citep{kahn1962topological}. Of course, in causal discovery, we are not given the graph, so implementing SOUR/SIR is not straightforward. In this paper, we propose a new method for causal discovery of layerings of discrete random variables, which implements SOUR/SIR without direct access to the graph, but with access to a conditional entropy oracle for the data instead. We show that, under some assumptions, we can separate sources from non-sources and sinks from non-sinks by comparing their conditional entropy (with appropriate conditioning) to their unconditional noise entropy. Specifically, when repeatedly removing sources and conditioning on all removed variables, we show that the conditional entropy of new sources equals the conditional entropy of their noise, while the conditional entropy of non-sources is larger than the entropy of their noise. On the other hand, when repeatedly removing sinks and conditioning on all non-removed variables, we show that the conditional entropy of new sinks equals the conditional entropy of their noise, while the conditional entropy of non-sinks is smaller than the entropy of their noise. Under our assumptions, our algorithms are provably correct and have a polynomial (quadratic) worst-case running-time.

We divide our assumptions to two: (i) assumptions not on noise entropies and (ii) assumptions on noise entropies. (i) includes faithfulness and injective noise (weaker than additive noise), while (ii) requires either that the noise entropies are known to us in advance, or---if the noise entropies are unknown---that the noise entropies are weakly monotonically increasing along directed paths. In addition, we require one of either a very mild extension of faithfulness, or strictly monotonically increasing noise entropies, or expanding noise injectivity to include an additional single argument in the structural functions. The known noise entropy assumption is of course much weaker than assuming known noise distribution. The weak monotonicity assumption is much weaker than the i.i.d. noise assumption, which received some attention in literature by \citet{xie2019efficient}, where the authors provide several references to real-world scenarios where the i.i.d. noise assumption holds. The assumptions we are not making are as important as those made: unlike many other existing methods of causal discovery, we are making no assumption about the topology of the causal graph such as sparsity, we do not assume any particular functional form such as linearity, and we do not assume any particular noise distribution.

The rest of the paper is organized as follows. Section \ref{subsec:literature} reviews related literature. Section \ref{sec:prelims} provides necessary definitions and preliminaries. Section \ref{sec:assumptions} collates all of the assumptions made in the paper. Section \ref{sec:bounds} shows how to bound the conditional entropy of a variable with its unconditional noise entropy. Section \ref{sec:algorithms} translates those bounds into causal discovery implementations of SOUR and SIR. Section \ref{sec:conclusion} concludes.

\subsection{Related Literature}\label{subsec:literature}

Causal discovery is a rich and evolving field. Several detailed surveys of causal discovery exist \citep{zanga2022survey, vowels2022d, nogueira2022methods, spirtes2016causal}, as well as a book on the subject \citep{peters2017elements}. As it is known that full causal discovery becomes much easier given a layering, finding such a layering plays a key role in many papers  \citep{ruiz2022sequentially, sanchez2022diffusion, montagna2023causal, teyssier2005ordering}. Information theory has been used in causal discovery for a variety of purposes \citep{kocaoglu2020applications, branchini2023causal,cabelireliable,runge2018conditional,marx2021information}.

A somewhat related approach to ours is entropic causal inference (ECI) \citep{compton2022entropic}. Like our method, ECI also implements SOUR by considering noise entropy, but in a very different way. First, the main assumption ECI makes on noise entropy is that it is small, as opposed to our assumption of either known or monotonic noise entropy. Second, ECI assumes access to both a conditional independence oracle and a minimum entropy coupling oracle, unlike our method which assumes access to a conditional entropy oracle. An interesting similarity is that ECI's conditional independence testing uses the same conditioning set we use for conditional entropy in SOUR. Another somewhat related work is by \citet{xie2019efficient}, which studies the much more specific case of a linear model with i.i.d. noise. They too identify sources in SOUR using entropy minimization. Their method seems very different than ours, but it is possible---although we haven't been able to determine this---that it is a special case of our more general method (applied to linear model with i.i.d. noise).

\section{Preliminaries}\label{sec:prelims}

\subsection{Graph Theory}\label{subsec:graphtheory}

Before we move to causal discovery, let us begin with the realm of graph theory, where the graph is known. For a DAG $G=(V,E)$, we denote the edge $(v_i,v_j)$ as $v_i \rightarrow v_j$ (or equivalently $v_j \leftarrow v_i$). Note that in a DAG, at most one of the edges $v_i \rightarrow v_j$ and $v_i \leftarrow v_j$ can exist: we write $v_i-v_j$ when we want to refer to one of the edges $v_i \rightarrow v_j$ or $v_i \leftarrow v_j$ but not specify the direction, so the statement $v_i - v_j \in E$ should be read as ``$v_i \rightarrow v_j \in E$ or $v_i \leftarrow v_j \in E$". For $v \in V$, we define $Par(v,G)=\{v' \in V: v' \rightarrow v \in E\}$ as the set of parents of $v$, and $Des(v,G)=\{v' \in V: \text{$v' \neq v$ and $\exists$ a directed path from $v$ to $v'$ in $G$}\}$ as the set of descendants of $v$. For any subsets $X,Y,S \subseteq V$, we denote as $X \vdash_{G} Y |S$ the event where $X$ and $Y$ are d-separated in $G$ conditional on $S$ (for information about d-separation, see \citet{geiger1990d}). For any $\widehat{V} \subseteq V$, we define the residual graph w.r.t. $\widehat{V}$ as $G^{\widehat{V}}=(\widehat{V},\{v \rightarrow v' \in E: v \in \widehat{V}\text{ and } v' \in \widehat{V}\})$ as the subgraph of $G$ corresponding to the nodes in $\widehat{V}$ and the edges between them. For a node $v \in V$, a parent $v' \in Par(v)$ that doesn't have any other parent as a descendant $Des(v') \cap Par(v) = \emptyset$ is called an {\it unmediated parent} of $v$. Nodes with only outgoing or only incoming edges are called sources and and sinks, respectively:
\begin{definitionsepcount}[sources and sinks]\label{def:sourcessinks}
Let $G=(V,E)$ be a digraph. $v \in V$ is called a source in $G$ if $v$ has no incoming edges, and a sink if $v$ has no outgoing edges. We denote the sets of all sources and sinks in $G$ respectively as $SRC(G)$ and $SNK(G)$.
\end{definitionsepcount}

Next, we define layerings \citep{tamassia2013handbook}, which are simply DAG topological orderings that allow for unbroken ties:
\begin{definitionsepcount}[layerings]\label{def:layerings}
Let $G=(V,E)$ be a digraph. Let $L=(L_1,\ldots,L_m)$ be a tuple of mutually exclusive non-empty subsets of $V$ ($L_i \cap L_j =\emptyset$ whenever $i \neq j$); we slightly abuse notation and overload $L$ to also mean $L=\cup_{i=1}^m{L_i}$ when clear from context. We write $L_i < L_j$ iff $i<j$ (and similarly define $>,=,\leq,\geq$). For every $v \in V$, if $v \in L_i$, define $L(v)=L_i$. $L$ is a {\it layering} iff it is a partition of $V$ and for all $v, v' \in V$, $v \rightarrow v' \in E \Rightarrow L(v) < L(v')$.
\end{definitionsepcount}
A digraph has a layering iff it is a DAG. Repeated removal of sources/sinks yields a layering; we precisely define this process in Algorithm \ref{alg:rr}, which we call the {\it Repeated Removal (RR) Algorithm}. This is a simple generalized variant of the well-known Kahn's Algorithm \citep{kahn1962topological}, but we include a correctness proof here for completeness.\footnote{The literature we found technically describes special cases of RR, but their correctness proofs very easily extend to it; we provide this extension here for the reader's convenience. We believe it is highly likely that RR as we wrote it already exists in literature, but we could not find a reference.}
\begin{algorithm}
\caption{RepeatedRemoval (RR)}\label{alg:rr}
\begin{algorithmic}[1]
\REQUIRE DAG $G=(V,E)$
\ENSURE Layering $L$ of $G$
\STATE $V_{cur} \gets V$ 
\STATE $L_{start} \gets \text{empty sequence}$ 
\STATE $L_{end} \gets \text{empty sequence}$ 
 \WHILE{$V_{cur} \neq \emptyset$}
 \STATE $SR, SN \gets$ subsets of $SRC(G^{V_{cur}}), SNK(G^{V_{cur}})$ s.t. $SR \cup SN \neq \emptyset$
 \STATE $L_{start}.append(SR)$
 \STATE $L_{end}.prepend(SN)$
 \STATE $V_{cur} \gets V_{cur}-(SR \cup SN)$
 \ENDWHILE
 \STATE $L \gets concatenate(L_{start},L_{end})$
 \RETURN $L$
\end{algorithmic}
\end{algorithm}
\begin{theoremsepcount}[correctness of RR]\label{thm:rrcorrectness}
RR outputs a layering of its input (regardless of the specific choices made for $SR$ and $SN$ in each iteration).
\end{theoremsepcount}
\begin{proof}
Suppose $v \rightarrow v' \in E$. We need to show that $L(v)<L(v')$. We make the following useful observation: as long as $v, v' \in V_{cur}$, $v \notin SN$ and $v' \notin SR$. We break into cases:
\begin{enumerate}
\item Assume $v \in L_{start}$. If $v' \in L_{end}$, the result is trivial. If $v' \in L_{start}$, then our observation implies that $v'$ must have been {\bf appended} to $L_{start}$ at a strictly later iteration than $v$. Therefore, $v'$ is appended to $L_{start}$ when $v$ is already in $L_{start}$, implying $L_{start}(v)<L_{start}(v')$.
\item Assume $v \in L_{end}$. Then our observation implies that $v' \in L_{end}$, and furthermore that $v$ must be added to $L_{end}$ at a strictly later iteration than $v'$. Therefore, $v$ is {\bf prepended} to $L_{end}$ when $v'$ is already in $L_{end}$, implying $L_{end}(v)<L_{end}(v')$.
\end{enumerate}
\end{proof}
When RR is implemented with $SN=\emptyset$ at all times (that is, in every iteration only sources are removed, but not necessarily all sources), we refer to it as the {\it repeatedSOUrceRemoval (SOUR) Algorithm}. Alternatively, when RR is implemented with $SR=\emptyset$, we refer to it as the {\it repeatedSInkRemoval (SIR) Algorithm}. We provide the simplified pseudocode for SOUR and SIR as Algorithms \ref{alg:sour} and \ref{alg:sir} respectively.

\noindent\begin{minipage}{0.5\textwidth}
\begin{algorithm}[H]
\caption{repeatedSOUrceRemoval (SOUR)}\label{alg:sour}
\begin{algorithmic}[1]
\REQUIRE DAG $G=(V,E)$
\ENSURE Layering $L$ of $G$
\STATE $V_{cur} \gets V$
\STATE $L \gets \text{empty sequence}$
 \WHILE{$V_{cur} \neq \emptyset$}
 \STATE $SR \gets$ non-$\emptyset$ subset of $SRC(G^{V_{cur}})$
 \STATE $L.append(SR)$
 \STATE $V_{cur} \gets V_{cur}-SR$
 \ENDWHILE
 \RETURN $L$
\end{algorithmic}
\end{algorithm}
\end{minipage}
\begin{minipage}{0.5\textwidth}
\begin{algorithm}[H]
\caption{repeatedSInkRemoval (SIR)}\label{alg:sir}
\begin{algorithmic}[1]
\REQUIRE DAG $G=(V,E)$
\ENSURE Layering $L$ of $G$
\STATE $V_{cur} \gets V$
\STATE $L \gets \text{empty sequence}$
 \WHILE{$V_{cur} \neq \emptyset$}
 \STATE $SN \gets$ non-$\emptyset$ subset of $SNK(G^{V_{cur}})$
 \STATE $L.prepend(SN)$
 \STATE $V_{cur} \gets V_{cur}-SN$
 \ENDWHILE
 \RETURN $L$
\end{algorithmic}
\end{algorithm}
\end{minipage}

\subsection{Causality}\label{subsec:causality}

Our goal in this paper is to recover a causal DAG's layering without knowing the graph, but with access---via a conditional entropy oracle---to data generated by it. As we will show, under some assumptions, SOUR and SIR can be implemented in this scenario. Throughout this paper, let $G_c=(V_c,E_c)$ be a DAG, which we refer to as the {\it causal graph}. We slightly abuse notation and consider the nodes in $V$ also as random variables. We assume that the variables in $V$ are connected through a structural causal model (SCM), meaning that for each $v \in V$, $v=f_v(Par(v,G_c) \cup \{N_v\})$ for some function $f_v$ and noise variable $N_v$ independent of all other noise variables. To clarify, $f_v(Par(v,G_c) \cup \{N_v\})$ is also a slight abuse of notation, which when $Par(v,G_c)=\{p_1,\ldots,p_t\}$ means $f_v(p_1,\ldots,p_t,N_v)$ (using the appropriate order of arguments). We assume that all noise variables (and hence all variables) are discrete, so the definitions of entropy and conditional entropy apply.

$G_c$ is unknown to us, but we assume access to a conditional entropy oracle $\mathscr{H}$. For sets of random variables $X$, $Y$ and $S$, let $\mathscr{H}(X|S)$ be the conditional entropy of $X$ conditional on $S$. We denote as $X \indep Y |S$ the case where $X$ and $Y$ are independent conditional on $S$. In the notations $\mathscr{H}(X|S)$, $X \indep Y |S$, and $X \vdash_{G} Y |S$, we allow replacing singleton sets with their element (e.g. if $X=\{x\}$ then $\mathscr{H}(x|S)=\mathscr{H}(X|S)$ etc.), and when $S=\emptyset$ we allow dropping it from the notation (e.g. $\mathscr{H}(X)=\mathscr{H}(X|\emptyset)$ etc.). Whenever we drop the graph from notation, the underlying graph we refer to is $G_c$ (e.g. we allow writing $Par(v)$ instead of $Par(v,G_c)$, $\vdash$ instead of $\vdash_{G_c}$, etc.). We also define the explicit noise graph, which makes the noise terms into explicit nodes:
\begin{definitionsepcount}[explicit noise graph]\label{def:explicit}
The {\it explicit noise graph} $G^N_c=(V^N_c,E^N_c)$ is obtained from $G_c$ by adding a node $N_v$ for each $v \in V_c$, with exactly one adjacent edge $N_v \rightarrow v$. That is, $V^N_c = V_c \cup \{N_v : v \in V_c\}$ and $E^N_c = E_c \cup \{N_v \rightarrow v: v \in V_c\}$.
\end{definitionsepcount}
We point out that d-separation implies conditional independence \citep{Pearl00}:
\begin{theoremsepcount}\label{thm:dsepimpliesind}
For $X,Y,S \subseteq V_c$, $X \vdash_{G_c} Y |S \Rightarrow X \indep Y |S$ and similarly for every $X,Y,S \subseteq V_c^N$, $X \vdash_{G_c^N} Y |S \Rightarrow X \indep Y |S$.
\end{theoremsepcount}

\section{Assumptions}\label{sec:assumptions}

Our layering discovery method requires some assumptions, which we list here. We break our assumptions to three sets. Assumption Collection \ref{assum:global} contains {\it global} assumptions, which we implicitly make throughout the rest of the paper from this point on. Assumption Collections \ref{assum:nonnoiseentropy} and \ref{assum:noiseentropy} contain {\it local} assumptions, which are {\bf not implicitly assumed} but instead are only assumed when explicitly stated; we collate them here as a convenient reference point.  Let us first introduce our global assumptions:
\begin{assumptioncollection}[global assumptions]\label{assum:global}
We always make the following assumptions.
\begin{enumerate}[label={\theassumptioncollection.\arabic*}]
\item \underline{\it Faithfulness:} For $X,Y,S \subseteq V_c$, $X \indep Y |S \Rightarrow X \vdash Y |S$. \label{assum:faithfulness}
\item \underline{\it Injective noise:} For all $v \in V_c$, when holding the values of $Par(v)$ constant, $f_v$ as a function of $N_v$ is one-to-one. \label{assum:injectivenoise}
\item \underline{\it Non-constant noise:} For every $v \in V_c$, $N_v$ is not constant, or equivalently $\mathscr{H}(N_v) > 0$. \label{assum:nonconstantnoise}
\end{enumerate}
\end{assumptioncollection}
Assumption \ref{assum:faithfulness} is common throughout the causal discovery literature \citep{spirtes2000causation}, Assumption \ref{assum:injectivenoise} is weaker than the Additive Noise Assumption \citep{hoyer2008nonlinear}, and Assumption \ref{assum:nonconstantnoise} simply prevents some degenerate cases. Next, we state two local assumptions, not about noise entropy, which are used in different parts of Theorem \ref{thm:conditionalentropy} in Section \ref{sec:bounds} to bound the conditional entropy of variables by the entropy of their noise.
\begin{assumptioncollection}[non noise entropy local assumptions]\label{assum:nonnoiseentropy}
We occasionally make some of the following assumptions.
\begin{enumerate}[label={\theassumptioncollection.\arabic*}]
\item \underline{\it Injective noise plus one:} For all $v \in V$, $v' \in Par(v)$, when holding the values of $Par(v)-\{v'\}$ constant, $f_v$ as a function of $v'$ and $N_v$ is one-to-one. \label{assum:injectivenoiseplusone}
\item \underline{\it Directed-faithfulness:} For every $v \in V$, $v' \in Des(v)$,  $N_v \cancel{\indep} v'$. \label{assum:directedfaithfulness}
\end{enumerate}
\end{assumptioncollection}
Assumption \ref{assum:injectivenoiseplusone} implies Assumption \ref{assum:injectivenoise}, and Assumption \ref{assum:directedfaithfulness} is weaker than faithfulness on the explicit noise graph. Finally, we state another list of local assumptions, about noise entropy, which are used for our implementation of SOUR and SIR via a conditional entropy oracle in Section \ref{sec:algorithms}. These assumptions allow us to use the bounds established by Theorem \ref{thm:conditionalentropy} to extract layering information.
\begin{assumptioncollection}[noise entropy local assumptions]\label{assum:noiseentropy}
We occasionally make some of the following assumptions.
\begin{enumerate}[label={\theassumptioncollection.\arabic*}]
\item \underline{\it Known noise entropy:} For all $v \in V$, $\mathscr{H}(N_v)$ is known to us.\label{assum:knownnoiseentropy}
\item \underline{\it Weakly increasing noise entropy:} For all $v \in V$, $v' \in Des(v)$, $\mathscr{H}(N_v) \leq \mathscr{H}(N_{v'})$. \label{assum:weakincreasingnoiseentropy}
\item \underline{\it Strictly increasing noise entropy:} For all $v \in V$, $v' \in Des(v)$, $\mathscr{H}(N_v) < \mathscr{H}(N_{v'})$. \label{assum:strictincreasingnoiseentropy}
\end{enumerate}
\end{assumptioncollection}
Assumption \ref{assum:knownnoiseentropy} is clearly much weaker than assuming the noise distribution is known in advance. Assumption \ref{assum:weakincreasingnoiseentropy} is satisfied in the case of i.i.d. noise, but is of course a weaker requirement than i.i.d. noise. The i.i.d. noise case received some attention in literature by \citet{xie2019efficient}, who provide references to real-world scenarios where the i.i.d. noise assumption holds. Assumption \ref{assum:strictincreasingnoiseentropy} is a bit stronger than \ref{assum:weakincreasingnoiseentropy}, and rules out the i.i.d. noise case.

\section{Entropy Bounds}\label{sec:bounds}

Let us first provide some informal intuition for our bounds. The value of $v \in V_c$ is fully determined by $Par(v)$ and $N_v$. When we condition on $Par(v)$, $N_v$ remains the only source of randomness for $v$. If we don't condition on anything else, then since noise injectivity implies that $f_v$ doesn't dilute $N_v$'s entropy, we get that the conditional entropy of $v$ equals $\mathscr{H}(N_v)$. If the conditioning set includes---in addition to $Par(v)$---some descendant $u \in Des(v)$, then $u$ generally carries additional information about $N_v$ and we can expect the entropy to be reduced relatively to $\mathscr{H}(N_v)$. On the other hand, if the conditioning set excludes some unmediated parent $r \in Par(v)$, then $r$ is a source of randomness additional to $N_v$. If the conditioning set also excludes all descendants of $r$ (and thus also all descendants of $v$), then no additional information is given about $N_v$ or $r$, so (using noise-plus-one injectivity to prevent $f_v$ from diluting the entropy) we can generally expect a conditional entropy larger than $\mathscr{H}(N_v)$.

Theorem \ref{thm:conditionalentropy} formalizes the intuition above. In Section \ref{sec:algorithms}, we use it to detect sinks and sources in graphs via a conditional entropy oracle. First, we introduce a lemma:
\begin{lemmasepcount}\label{lem:dsep}
Let $v \in V_c$ and $S \subseteq V_c-\{v\}$ s.t. $Des(v,G_c) \cap S = \emptyset$. Then $N_v \vdash_{G^N_c} S$, and therefore $N_v \indep S$.
\end{lemmasepcount}
\begin{proof}
Consider any path of undirected edges $N_v = u_0 - u_1 - u_2 - \cdots u_k = s$ between $N_v$ and a node $s \in S$. The adjacent edge to $N_v$ must necessarily be the edge $N_v \rightarrow v$, since this is the only edge adjacent to $N_v$ in the graph; in particular, $u_1=v$. Since $s$ is not a descendant of $v$, the remainder of the path $u_1 = v - \cdots - s = u_k$ must contain at least one edge of the form $u_{i} \leftarrow u_{i+1}$; choose $i$ to be the minimum value so that  $u_{i} \leftarrow u_{i+1}$ exists in the path. Since we have established $i \geq 1$, then the previous edge $u_{i-1}-u_i$ exists and must be oriented as $u_{i-1} \rightarrow u_i$; therefore, $u_i$ is a collider in the path, and since we are not conditioning on anything, neither $u_i$ nor any of its descendants are conditioned on. Therefore, we have shown that $N_v \vdash_{G^N_c} S$, and therefore $ N_v \indep S$ (by Theorem \ref{thm:dsepimpliesind}).
\end{proof}
We can now state and prove Theorem \ref{thm:conditionalentropy}.
\begin{theoremsepcount}[entropy bounds]\label{thm:conditionalentropy}
Let $v \in V_c$ and let $S \subseteq V-\{v\}$. Then:
\begin{enumerate}[label={\thetheoremsepcount.\arabic*}]
\item Assume $Par(v) \subseteq S$. Then $\mathscr{H}(v|S) \leq \mathscr{H}(N_v)$. (Conditioning on a node's parents yields a weakly lower entropy than the noise's.)\footnote{Theorem \ref{thm:weakupper} does not actually require Assumption \ref{assum:injectivenoise}, but we will only be using this part of the theorem in conjunction with Theorem \ref{thm:eq} which does require that assumption, so there is no need to separate the cases further.}\label{thm:weakupper}
\item Assume $Par(v) \subseteq S$ and $Des(v) \cap S = \emptyset$. Then $\mathscr{H}(v|S)=\mathscr{H}(N_v)$. (Conditioning on a node's parents but no descendants yields exactly the same entropy as the noise's.) \label{thm:eq}
\item Assume Assumption \ref{assum:directedfaithfulness}. Assume $Par(v) \subseteq S$ and $Des(v) \cap S \neq \emptyset$. Then $\mathscr{H}(v|S) < \mathscr{H}(N_v)$. (Conditioning on a node's parents and some descendants yields a strictly lower entropy than the noise's.)\label{thm:strictupper}
\item Assume Assumption \ref{assum:injectivenoiseplusone}. Assume that there exists $v' \in Par(v)-S$ s.t. $Des(v') \cap (S \cup Par(v))=\emptyset$ . Then $\mathscr{H}(v|S) > \mathscr{H}(N_v)$. (Failing to condition on at least one unmediated parent of a node and that parent's descendants yields a strictly higher entropy than the noise's.)\label{thm:lower}
\end{enumerate}
\end{theoremsepcount}
\begin{proof}
\begin{enumerate}
\item[3.1] Compute:
\begin{align*}
&\mathscr{H}(v|S)\\
&=\mathscr{H}(f_v(Par(v) \cup \{N_v\})|S) \\
&\leq \mathscr{H}(Par(v) \cup \{N_v\}|S) & \\
&= \mathscr{H}(Par(v)|S)+\mathscr{H}(N_v|S \cup Par(v)) & \text{chain rule} \\
&= 0+\mathscr{H}(N_v|S) & Par(v) \subseteq S \\
&= \mathscr{H}(N_v|S) & \\
&\leq \mathscr{H}(N_v). &
\end{align*}
\item[3.2] Compute:
\begin{align*}
&\mathscr{H}(v|S)\\
&=\mathscr{H}(f_v(Par(v) \cup \{N_v\})|S) \\
&= \mathscr{H}(Par(v) \cup \{N_v\}|S) & \text{$Par(v) \subseteq S$ and Assumption \ref{assum:injectivenoise}} \\
&= \ldots = \mathscr{H}(N_v|S) & \text{as in the proof of Theorem \ref{thm:weakupper}}\\
&= \mathscr{H}(N_v). & \text{$N_v \indep S$ by Lemma \ref{lem:dsep}}
\end{align*}
\item[3.3] Compute:
Assumption \ref{assum:directedfaithfulness} implies that $N_v$ is dependent on $S$ (since $S$ contains at least one element from $Des(v)$), and therefore $\mathscr{H}(N_v|S)<\mathscr{H}(N_v)$. So now we have:
\begin{equation*}
\mathscr{H}(v|S) \leq \mathscr{H}(N_v|S)<\mathscr{H}(N_v),
\end{equation*}
where the first inequality is by the proof of Theorem \ref{thm:weakupper}.
\item[3.4]  We make two initial observations:
\begin{enumerate}
\item[(1)] $N_v \indep S \cup Par(v)$. This is because $Des(v') \cap S = \emptyset$ and $Des(v) \subset Des(v')$, so $Des(v) \cap S =\emptyset$ and trivially $Des(v) \cap Par(v)=\emptyset$. Thus Lemma \ref{lem:dsep} implies $N_v \indep S \cup Par(v)$.
\item[(2)] $N_{v'} \indep S \cup (Par(v)-\{v'\}) \cup Par(v')$. This is because $Des(v') \cap (S \cup (Par(v)-\{v'\})) = \emptyset$ by assumption, and trivially $Des(v') \cap Par(v') =\emptyset$, so Lemma \ref{lem:dsep} implies $N_{v'} \indep S \cup (Par(v)-\{v'\}) \cup Par(v')$.
\end{enumerate}
 Compute:
\begin{align*}
&\mathscr{H}(v|S)\\
&=\mathscr{H}(f_v(Par(v) \cup \{N_v\})|S)\\
&=\mathscr{H}(f_v((Par(v)-\{v'\}) \cup \{N_v, v'\})|S) \\
&\geq \mathscr{H}(f_v((Par(v)-\{v'\}) \cup \{N_v, v'\})|S \cup (Par(v)-\{v'\})) \\
& = \mathscr{H}(N_v, v'|S \cup (Par(v)-\{v'\})) & \text{Assumption \ref{assum:injectivenoiseplusone}}\\
&=\mathscr{H}(N_v|S \cup Par(v))+\mathscr{H}(v'|S \cup (Par(v)-\{v'\})) & \text{chain rule} \\
&=\mathscr{H}(N_v)+\mathscr{H}(v'|S \cup (Par(v)-\{v'\})) & \text{Observation (1)}\\
\end{align*}
To complete our proof, it is sufficient to show that $\mathscr{H}(v'|S \cup (Par(v)-\{v'\}))>0$:
\begin{align*}
&\mathscr{H}(v'|S \cup (Par(v)-\{v'\}))\\
&=\mathscr{H}(f_{v'}(Par(v')\cup \{N_{v'}\})|S \cup (Par(v)-\{v'\}))\\
& \geq \mathscr{H}(f_{v'}(Par(v')\cup \{N_{v'}\})|S \cup (Par(v)-\{v'\}) \cup Par(v'))\\
& =  \mathscr{H}(N_{v'}|S \cup (Par(v)-\{v'\}) \cup Par(v')) & \text{Assumption \ref{assum:injectivenoise}}\\
& = \mathscr{H}(N_{v'}) & \text{Observation (2)}\\
& > 0 & \text{Assumption \ref{assum:nonconstantnoise}}\\
\end{align*}
\end{enumerate}
\end{proof}

\section{Causal Layering Algorithms}\label{sec:algorithms}

In this section, we use Theorem \ref{thm:conditionalentropy} to implement SOUR and SIR when we can only access the graph via a conditional entropy oracle. Looking at the psuedocode of SOUR in Algorithm \ref{alg:sour} (resp. SIR in Algorithm \ref{alg:sir}), we see that information about edges is only used in line 4, to identify a non-empty subset of sources (resp. sinks). With certain assumptions and conditioning, sources' (resp. sinks') conditional entropy is equal to their noise entropy, while the conditional entropy of non-sources (resp. non-sinks) is larger (resp. smaller) than their noise entropy. We can use this separation to implement line 4 without knowledge of the underlying graph, but with a conditional entropy oracle. SOUR can be implemented subject to Assumptions \ref{assum:injectivenoiseplusone} and either \ref{assum:knownnoiseentropy} or \ref{assum:weakincreasingnoiseentropy}. SIR can be implemented subject to Assumptions \ref{assum:directedfaithfulness} and either \ref{assum:knownnoiseentropy} or \ref{assum:weakincreasingnoiseentropy}. SIR can also be implemented subject just to Assumption \ref{assum:strictincreasingnoiseentropy}.

\thickmuskip=0.19\thickmuskip
\noindent\begin{minipage}{0.5\textwidth}
\begin{algorithm}[H]
\caption{SOUR (Causal Discovery)}\label{alg:sourcd}
\begin{algorithmic}[1]
\ASSUMPTIONS \ref{assum:injectivenoiseplusone} \& (\ref{assum:knownnoiseentropy} or \ref{assum:weakincreasingnoiseentropy})
\REQUIRE Variables $V_c$, entropy oracle $\mathscr{H}$
\ENSURE A Layering $L$ of $G_c$
\STATE $V_{cur} \gets V_c$
\STATE $L \gets \text{empty sequence}$
 \WHILE{$V_{cur} \neq \emptyset$}
\IF{As. \ref{assum:injectivenoiseplusone} \& \ref{assum:knownnoiseentropy} hold}
 \STATE $SR \gets$ non-$\emptyset$ subset of\\$\{v \in V_{cur}:\mathscr{H}(v|V_c-V_{cur})=\mathscr{H}(N_v)\}$
 \ELSIF{As. \ref{assum:injectivenoiseplusone} \& \ref{assum:weakincreasingnoiseentropy} hold}
 \STATE $SR \gets$ non-$\emptyset$ subset of\\$\argmin_{v \in V_{cur}}{\mathscr{H}(v|V_c-V_{cur})}$
 \ENDIF
 \STATE $L.append(SR)$
 \STATE $V_{cur} \gets V_{cur}-SR$
 \ENDWHILE
 \RETURN $L$
\end{algorithmic}
\end{algorithm}
\end{minipage}
\begin{minipage}{0.5\textwidth}
\begin{algorithm}[H]
\caption{SIR (Causal Discovery)}\label{alg:sircd}
\begin{algorithmic}[1]
\ASSUMPTIONS \ref{assum:directedfaithfulness} \& (\ref{assum:knownnoiseentropy} or \ref{assum:weakincreasingnoiseentropy}), or \ref{assum:strictincreasingnoiseentropy}
\REQUIRE Variables $V_c$, entropy oracle $\mathscr{H}$
\ENSURE A Layering $L$ of $G_c$
\STATE $V_{cur} \gets V_c$
\STATE $L \gets \text{empty sequence}$
 \WHILE{$V_{cur} \neq \emptyset$}
\IF{As. \ref{assum:directedfaithfulness} \& \ref{assum:knownnoiseentropy} hold}
 \STATE $SN \gets$ non-$\emptyset$ subset of\\$\{v \in V_{cur}:\mathscr{H}(v|V_{cur}-\{v\})=\mathscr{H}(N_v)\}$
 \ELSIF{As. \ref{assum:directedfaithfulness} \& \ref{assum:weakincreasingnoiseentropy} or As. \ref{assum:strictincreasingnoiseentropy} hold}
 \STATE $SN \gets$ non-$\emptyset$ subset of\\$\argmax_{v \in V_{cur}}{\mathscr{H}(v|V_{cur}-\{v\})}$
 \ENDIF
 \STATE $L.prepend(SN)$
 \STATE $V_{cur} \gets V_{cur}-SN$
 \ENDWHILE
 \RETURN $L$
\end{algorithmic}
\end{algorithm}
\end{minipage}
\thickmuskip=5.26315789474\thickmuskip
Algorithms \ref{alg:sourcd} and \ref{alg:sircd} present the causal discovery implementations of SOUR and SIR respectively. By $\argmin$/$\argmax$ we mean the set of all minimizing/maximizing arguments and not just an arbitrary one (in case the minimizing/maximizing argument is unique, then $\argmin$/$\argmax$ is a singleton set). Lines 4-8 in Algorithms \ref{alg:sourcd} and \ref{alg:sircd} replace line 4 in Algorithms \ref{alg:sour} and \ref{alg:sir}. Therefore, to guarantee the correctness of the causal discovery implementations, we must show  that the replacement lines accomplish the same function as the original line 4. Specifically, we need to show that the replacement lines produce a non-empty subset of $SRC(G_c^{V_{cur}})$ for SOUR and $SNK(G_c^{V_{cur}})$ for SIR. This is where Theorem \ref{thm:conditionalentropy} comes in handy. Theorem \ref{thm:cdcorrectness} proves that Algorithms \ref{alg:sourcd} and \ref{alg:sircd} implement SOUR and SIR (Algorithms \ref{alg:sour} and \ref{alg:sir} respectively), thus---by Theorem \ref{thm:rrcorrectness}---Algorithms \ref{alg:sourcd} and \ref{alg:sircd} produce a layering of $G_c$:
\begin{theoremsepcount}[correctness of Algorithms \ref{alg:sourcd} and \ref{alg:sircd}]\label{thm:cdcorrectness}
The following statements hold whenever the relevant algorithm reaches line 4. By SOUR we refer to Algorithm \ref{alg:sourcd} and by SIR we refer to Algorithm \ref{alg:sircd}.
\begin{enumerate}[label={\thetheoremsepcount.\arabic*}]
\item Assume As. \ref{assum:injectivenoiseplusone}. In SOUR, $SRC(G_c^{V_{cur}})=\{v \in V_{cur}:\mathscr{H}(v|V_c-V_{cur})=\mathscr{H}(N_v)\}$. \label{thm:sourknown}
\item Assume As. \ref{assum:injectivenoiseplusone} \& \ref{assum:weakincreasingnoiseentropy}. In SOUR, $\argmin_{v \in V_{cur}}{\mathscr{H}(v|V_c-V_{cur})} \subseteq SRC(G_c^{V_{cur}})$. \label{thm:sourunknown}
\item Assume As. \ref{assum:directedfaithfulness}. In SIR, $SNK(G_c^{V_{cur}})=\{v \in V_{cur}:\mathscr{H}(v|V_{cur}-\{v\})=\mathscr{H}(N_v)\}$. \label{thm:sirknown}
\item Assume As. \ref{assum:directedfaithfulness} \& \ref{assum:weakincreasingnoiseentropy}, or \ref{assum:strictincreasingnoiseentropy}. In SIR, $\argmax_{v \in V_{cur}}{\mathscr{H}(v|V_{cur}-\{v\})} \subseteq SNK(G_c^{V_{cur}})$. \label{thm:sirunknown}
\end{enumerate}
\end{theoremsepcount}
\begin{proof}
Note that $V_{cur}$ is monotonically shrinking throughout iterations in both SOUR and SIR. For each proof, we assume the algorithm worked correctly up until the current iteration: that is, in all previous iterations,  $SR \subseteq SRC(G_c^{V_{cur}})$ for SOUR and  $SN \subseteq SNK(G_c^{V_{cur}})$ for SIR (where $V_{cur}$ is the value of $V_{cur}$ at that iteration). That is, SOUR only removed sources and SIR only removed sinks from the residual graph of each previous iteration.
\begin{enumerate}
\item[4.1] Let $v \in V_{cur}$. Consider $v \in SRC(G_c^{V_{cur}})$. Since $v$ is a source, all nodes in $Par(v,G_c)$ are no longer in $V_{cur}$, and thus $Par(v,G_c) \subseteq V_c-V_{cur}$. On the other hand, since $v \in V_{cur}$, $v$ has never been removed and so the nodes in $Des(v,G_c)$ were never sources in previous iterations, and therefore $Des(v,G_c) \subseteq V_{cur}$, thus $Des(v,G_c) \cap (V_c-V_{cur}) =\emptyset$. Theorem \ref{thm:eq} therefore implies that $\mathscr{H}(v|V_c-V_{cur})=\mathscr{H}(N_v)$.

Consider instead $v \notin SRC(G_c^{V_{cur}})$. Since $v \notin SRC(G_c^{V_{cur}})$, then $Par(v,G_c) \cap V_{cur} \neq \emptyset$, so there exists some $v^* \in Par(v,G_c) \cap V_{cur}$. We claim that there exists an unmediated parent $v' \in Par(v,G_c) \cap V_{cur}$ of $v$, meaning that $Des(v',G_c) \cap Par(v,G_c) = \emptyset$. We will construct a finite sequence $u_0,u_1,u_2,\ldots,u_k$ which satisfies the following properties:
\begin{itemize}
\item $u_0=v^*$
\item $u_{i+1} \in Des(u_i,G_c)$ for all $i$
\item $u_i \in Par(v,G_c) \cap V_{cur}$ for all $i$
\item $Des(u_k,G_c) \cap Par(v,G_c) = \emptyset$
\end{itemize}
Once we construct this sequence, we can set $v'=u_k$ and our claim is proven. As $v'$ is an ancestor of all nodes in $Des(v',G_c)$ and $v' \in V_{cur}$, then none of the nodes in $Des(v',G_c)$ have been removed either (they were never sources since they have an unremoved ancestor), meaning $Des(v',G_c) \subseteq V_{cur}$ so $Des(v',G_c) \cap (V_c-V_{cur})=\emptyset$. Since also $v' \in Par(v,G_c) \cap V_{cur}$ and $Des(v',G_c) \cap Par(v,G_c) = \emptyset$, we can apply Theorem \ref{thm:lower} to establish $\mathscr{H}(v|V_c-V_{cur})>\mathscr{H}(N_v)$.

All that is left is to construct the sequence. For every $i$, if $Des(u_i,G_c) \cap Par(v,G_c) = \emptyset$, we can simply set $k=i$ and end the sequence. If $Des(u_i,G_c) \cap Par(v,G_c) \neq \emptyset$, then there exists some $u_{i+1} \in Des(u_i,G_c) \cap Par(v,G_c)$; furthermore, since $u_i \in V_{cur}$ and $u_i$ is an ancestor of $u_{i+1}$, it follows that $u_{i+1}$ was never removed from $V_{cur}$ either, and therefore $u_{i+1} \in Par(v,G_c) \cap V_{cur}$. Since each element of the sequence is a descendant of the previous one, the sequence is moving down a directed path (potentially skipping some nodes along the path), and since $G_c$ is acyclic, this means that the sequence contains no repetitions. As the $V_c$ is finite, the sequence must be finite, and therefore it must end with some $u_i$ satisfying $Des(u_i,G_c) \cap Par(v,G_c) = \emptyset$, so $k$ must be finite.

\item[4.2] Let $u \in V_{cur}$. Our proof of Theorem \ref{thm:sourknown} established that if $q \in SRC(G_c^{V_{cur}})$, then $\mathscr{H}(q|V_c-V_{cur})=\mathscr{H}(N_q)$, and if $q \notin SRC(G_c^{V_{cur}})$, then $\mathscr{H}(q|V_c-V_{cur})>\mathscr{H}(N_q)$. Assume $u \notin SRC(G_c^{V_{cur}})$. In that case, $\mathscr{H}(u|V_c-V_{cur})>\mathscr{H}(N_{u})$; also, $u$ has at least one ancestor $u' \in SRC(G_c^{V_{cur}})$, and for that ancestor $\mathscr{H}(u'|V_c-V_{cur})=\mathscr{H}(N_{u'})$. Assumption \ref{assum:weakincreasingnoiseentropy} implies $\mathscr{H}(N_{u'}) \leq \mathscr{H}(N_u)$, and therefore we have 
\begin{equation*}
\mathscr{H}(u'|V_c-V_{cur})=\mathscr{H}(N_{u'}) \leq \mathscr{H}(N_u) < \mathscr{H}(u|V_c-V_{cur}).
\end{equation*}
Thus, $u \notin \argmin_{v \in V_{cur}}{\mathscr{H}(v|V_c-V_{cur})}$.

\item[4.3] Let $v \in V_{cur}$. Since $v \in V_{cur}$, and $v$ is a child of all nodes in $Par(v,G_c)$, then no node in $Par(v,G_c)$ has been a sink in any previous iteration, and therefore $Par(v,G_c) \subseteq V_{cur}$, so $Par(v,G_c) \subseteq V_{cur}-\{v\}$. If $v \in SNK(G_c^{V_{cur}})$, then $Des(v,G_c) \cap V_{cur} =\emptyset$, and since also $Par(v,G_c) \subseteq V_{cur}-\{v\}$, Theorem \ref{thm:eq} implies that $\mathscr{H}(v|V_{cur}-\{v\})=\mathscr{H}(N_v)$. If instead $v \notin SNK(G_c^{V_{cur}})$, then $Des(v,G_c) \cap V_{cur} \neq \emptyset$, so by Theorem \ref{thm:strictupper} $\mathscr{H}(v|V_{cur}-\{v\})<\mathscr{H}(N_v)$.

\item[4.4] Let $u \in V_{cur}$. First, consider the case where Assumptions \ref{assum:directedfaithfulness} \& \ref{assum:weakincreasingnoiseentropy} hold. Because Assumption \ref{assum:directedfaithfulness} holds, our proof of Theorem \ref{thm:sirknown} established that if $q \in SNK(G_c^{V_{cur}})$, then $\mathscr{H}(q|V_{cur}-\{q\})=\mathscr{H}(N_q)$, and if $q \notin SNK(G_c^{V_{cur}})$, then $\mathscr{H}(q|V_{cur}-\{q\})<\mathscr{H}(N_q)$. Assume $u \notin SNK(G_c^{V_{cur}})$. In that case, $\mathscr{H}(u|V_{cur}-\{u\})<\mathscr{H}(N_u)$; also, $u$ has at least one descendant $u' \in SNK(G_c^{V_{cur}})$, and for that descendant $\mathscr{H}(u'|V_{cur}-\{u'\})=\mathscr{H}(N_{u'})$. Assumption \ref{assum:weakincreasingnoiseentropy} implies $\mathscr{H}(N_{u'}) \geq \mathscr{H}(N_u)$, and therefore we have
\begin{equation*}
\mathscr{H}(u'|V_{cur}-\{u'\})=\mathscr{H}(N_{u'}) \geq \mathscr{H}(N_u) >\mathscr{H}(u|V_{cur}-\{u\}).
\end{equation*}
Thus, $u \notin \argmax_{v \in V_{cur}}{\mathscr{H}(v|V_{cur}-\{v\})}$.

Alternatively, assume Assumption \ref{assum:strictincreasingnoiseentropy}. In the proof of Theorem \ref{thm:sirknown}, we have shown that if $q \in SNK(G_c^{V_{cur}})$ then $\mathscr{H}(q|V_{cur}-\{q\})=\mathscr{H}(N_q)$. That claim did not rely on Assumption \ref{assum:directedfaithfulness} and therefore still holds. On the other hand, since we only ever remove sinks, all the parents of nodes in $V_{cur}$ are also still in $V_{cur}$, and therefore Theorem \ref{thm:weakupper} implies that $\mathscr{H}(q|V_{cur}-\{q\}) \leq \mathscr{H}(N_q)$ for all $q \in V_{cur}$. Assume $u \notin SNK(G_c^{V_{cur}})$. As we have established, $\mathscr{H}(u|V_{cur}-\{u\}) \leq \mathscr{H}(N_u)$; also, $u$ has at least one descendant $u' \in SNK(G_c^{V_{cur}})$, and for that descendant $\mathscr{H}(u'|V_{cur}-\{u'\})=\mathscr{H}(N_{u'})$. Assumption \ref{assum:strictincreasingnoiseentropy} implies $\mathscr{H}(N_{u'}) > \mathscr{H}(N_u)$, and therefore we have
\begin{equation*}
\mathscr{H}(u'|V_{cur}-\{u'\})=\mathscr{H}(N_{u'}) > \mathscr{H}(N_u) \geq \mathscr{H}(u|V_{cur}-\{u\}).
\end{equation*}
Thus, $u \notin \argmax_{v \in V_{cur}}{\mathscr{H}(v|V_{cur}-\{v\})}$.
\end{enumerate}
\end{proof}
Finally, we present a straightforward observation regarding the running time of our algorithms. Note that the bound given is worst-case.
\begin{theoremsepcount}[running time of Algorithms \ref{alg:sourcd} and \ref{alg:sircd}]\label{thm:runtime}
Algorithms \ref{alg:sourcd} and \ref{alg:sircd} make $O(|V_c|^2)$ oracle calls.
\end{theoremsepcount}
\begin{proof}
In every iteration, the algorithms perform one oracle call for each element in $V_{cur}$. Furthermore, in the beginning $V_{cur}=V_c$, and in every iteration at least one element is removed from $V_{cur}$. Thus the number of oracle calls is bounded from above by $\sum_{i=1}^{|V_c|}{i}=O(|V_c|^2)$.
\end{proof}

\section{Conclusion}\label{sec:conclusion}

In this paper, we introduced a new class of causal discovery algorithms for discrete data. Our algorithms recover a layering of the causal graph, which they can only access via a conditional entropy oracle. In fact, our algorithms implement the SOUR and SIR algorithms from graph theory, but without direct access to the graph. The key idea behind our algorithms is that, with appropriate assumptions and conditioning, sources and sinks can be separated from the other nodes in the graph based on comparison between their conditional entropy and the unconditional entropy of their noise. The sources in SOUR and sinks in SIR have conditional entropy equal to the unconditional entropy of their noise. On the other hand, the non-sources in SOUR have conditional entropy larger than the unconditional entropy of their noise, while the non-sinks in SIR have conditional entropy smaller than the unconditional entropy of their noise. Our implementations of SOUR and SIR are provably correct and make $O(|V_c|^2)$ oracle calls in the worst-case. Our algorithms do not make many of the assumptions that are commonly made in literature, but they do need to make an assumption on noise entropies, namely that they are either known or monotonically increasing.
\bibliographystyle{plainnat}
\bibliography{layeringbibliography.bib}
\end{document}